\documentclass{article}
\usepackage[preprint, nonatbib]{neurips_2020}
\usepackage[utf8]{inputenc} 
\usepackage[T1]{fontenc}    
\usepackage{hyperref}       
\usepackage{url}            
\usepackage{booktabs}       
\usepackage{amsfonts}       
\usepackage{nicefrac}       
\usepackage{microtype}      
\usepackage{graphicx}
\usepackage{textcomp, setspace}
\usepackage{amsmath}
\title{Traffic flow prediction using Deep Sedenion Networks}

\author{
  Alabi Bojesomo, Hasan Al-Marzouqi, Panos Liatsis \\
  Electrical Engineering and Computer Science Department,\\
  Khalifa University, Abu Dhabi, UAE.\\
  \texttt{\{100046384, hasan.almarzouqi, panos.liatsis\}@ku.ac.ae} \\
}
\date{}  

\begin{document}

\maketitle

\begin{abstract}
   In this paper, we present our solution to the Traffic4cast2020 traffic prediction challenge. In this competition, participants are to predict future traffic parameters (speed and volume) in three different cities: Berlin, Istanbul and Moscow. The information provided includes nine channels where the first eight represent the speed and volume for four different direction of traffic (NE, NW, SE and SW), while the last channel is used to indicate presence of traffic incidents. The expected output should have the first 8 channels of the input at six future timing intervals (5, 10, 15, 30, 45, and 60min), while a one hour duration of past traffic data, in 5mins intervals, are provided as input. We solve the problem using a novel sedenion U-Net neural network. Sedenion networks provide the means for efficient encoding of correlated multimodal datasets. We use 12 of the 15 sedenion imaginary parts for the dynamic inputs and the real sedenion component is used for the static input. The sedenion output of the network is used to represent the multimodal traffic predictions. Proposed system achieved a validation MSE of 1.33e-3 and a test MSE of 1.31e-3.  

  
\end{abstract}

\section{Introduction}
\label{introduction}
In this contribution, we briefly summarize the methodology and experiments in tackling the Traffic4cast challenge 2020 \cite{traffic4cast}. The aim of the challenge is to predict future traffic flow volume and speed using traffic information projected on high resolution city maps. 
Given hourly traffic data, we need to build a model to predict traffic volume and speed at six time intervals: 5min, 10min, 15min, 30min, 45min and 1hr \cite{traffic4cast}.
The pixel level prediction of this task can be posed as a segmentation problem and hence, we used a U-Net \cite{unet} based architecture.\\ 
This year's competition includes both dynamic and static information, which are treated as independent modalities. Finding representations for multi-modal datasets can be done by using either early or late fusion. In late fusion, separate feature extracting networks are used to encode each individual modality. Encoded features are fused before the classification layer. Early fusion on the other hand has the modalities fused within the feature extraction step and before data encoding layers \cite{AUDEBERT201820_beyond_rgb}. The later approach is the one used in this paper as it did achieve higher levels of prediction accuracy when compared to late fusion architectures.  \\
To account for the multi-modal nature of the dataset, in the proposed U-Net model, a sedenion-based convolutional layer \cite{WU2020179_octonion, Saoud2020_sedenion} was used. Sedenion numbers are hypercomplex numbers composed from 16 dimensions and they provide efficient means for representing correlated multimodal datasets. The sedenion convolution layers are packed in blocks in a U-Net like architecture. The proposed model is highly parameter-efficient due to the use of sedenion convolutions and sedenion-based encoding of input and output parameters. 

\begin{figure}[tbp]
    \centering
    \includegraphics[width=1\columnwidth]{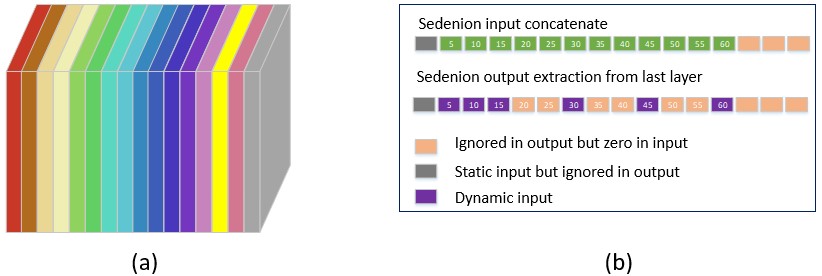}
    \caption{Sedenion input feature map division into 16 groups (a), and how the sedenion representation was formed in the first sedenion layer as well as how the outputs of the network was extracted from the last sedenion layer of the network (b)}
    \label{fig: sedenion divisions}
\end{figure}

Traffic static and dynamic information are represented using a 16 component sedenion representation. Since we have a single static frame and 12 dynamic frames as input parameters, we use the static frame as the first component of the input sedenion vector, while dynamic traffic information were represented using the next 12 components (Fig. \ref{fig: sedenion divisions}b). 

A sedenion can be defined as a 16-dimensional algebraic structure (Fig. \ref{fig: sedenion divisions}a), i.e.,  $X = x_0 + \sum_{k=1}^{15} x_k i_k$, where $x_0$ is the real part and $x_k, k\in(1,15)$ are the imaginary parts with $i_k^2 = -1$. Sedenion based convolution can be represented as the multiplication of two sedenion numbers \cite{WU2020179_octonion}. With weight $W = w_0 + \sum_{k=1}^{15} w_k i_k$ and input $X$ being a sedenion, the multiplication leads to a weight matrix having its component index and sign shown in equation (\ref{eqn: sedenion matrix}). The matrix shows that if $Y = W * X$ (eqn \ref{eqn: sedenion matrix eqn}), then $y_0 = w_0 x_0 - \sum_{k=1}^{15} w_k x_k$ following the first row of (\ref{eqn: sedenion matrix}). Other components of the output feature maps $y$ can be found accordingly. It is obvious that each of the 16 components of the sedenion weights are re-used 16 times, leading to parameter efficiency. Moreover, the parameter reduction does not come at the expense of representation ability because all the input feature map to the sedenion convolution takes part in the equation leading to each of the 16 sedenion outputs. To represent the inputs as a sedenion structure, the input feature map is divided into 16 groups, where each group represents a single component of the sedenion feature map.


{
\footnotesize
\begin{equation}
    \label{eqn: sedenion matrix}
    \begin{bmatrix}
           0  &  -1  &  -2  &  -3  &  -4  &  -5  &  -6  &  -7  &  -8  &  -9  & -10  & -11  & -12  &  -13  & -14  & -15 \\
          1  &   0  &  -3  &   2  &  -5  &   4  &   7  &  -6  &  -9  &   8  &  11  & -10  &  13  &
        -12  & -15  &  14 \\
          2  &   3  &   0  &  -1  &  -6  &  -7  &   4  &   5  & -10  & -11  &   8  &   9  &  14  &
         15  & -12  & -13 \\
          3  &  -2  &   1  &   0  &  -7  &   6  &  -5  &   4  & -11  &  10  &  -9  &   8  &  15  &
        -14  &  13  & -12 \\
          4  &   5  &   6  &   7  &   0  &  -1  &  -2  &  -3  & -12  & -13  & -14  & -15  &   8  &
          9  &  10  &  11 \\
          5  &  -4  &   7  &  -6  &   1  &   0  &   3  &  -2  & -13  &  12  & -15  &  14  &  -9  &
          8  & -11  &  10 \\
          6  &  -7  &  -4  &   5  &   2  &  -3  &   0  &   1  & -14  &  15  &  12  & -13  & -10  &
         11  &   8  &  -9 \\
          7  &   6  &  -5  &  -4  &   3  &   2  &  -1  &   0  & -15  & -14  &  13  &  12  & -11  &
        -10  &   9  &   8 \\
          8  &   9  &  10  &  11  &  12  &  13  &  14  &  15  &   0  &  -1  &  -2  &  -3  &  -4  &
         -5  &  -6  &  -7 \\
          9  &  -8  &  11  & -10  &  13  & -12  & -15  &  14  &   1  &   0  &   3  &  -2  &   5  &
         -4  &  -7  &   6 \\
         10  & -11  &  -8  &   9  &  14  &  15  & -12  & -13  &   2  &  -3  &   0  &   1  &   6  &
          7  &  -4  &  -5 \\
         11  &  10  &  -9  &  -8  &  15  & -14  &  13  & -12  &   3  &   2  &  -1  &   0  &   7  &
         -6  &   5  &  -4 \\
         12  & -13  & -14  & -15  &  -8  &   9  &  10  &  11  &   4  &  -5  &  -6  &  -7  &   0  &
          1  &   2  &   3 \\
         13  &  12  & -15  &  14  &  -9  &  -8  & -11  &  10  &   5  &   4  &  -7  &   6  &  -1  &
          0  &  -3  &   2 \\
         14  &  15  &  12  & -13  & -10  &  11  &  -8  &  -9  &   6  &   7  &   4  &  -5  &  -2  &
          3  &   0  &  -1 \\
         15  & -14  &  13  &  12  & -11  & -10  &   9  &  -8  &   7  &  -6  &   5  &   4  &  -3  &
         -2  &   1  &   0 \\
    \end{bmatrix}
\end{equation}
}

{
\small
\footnotesize
\begin{equation}
    \label{eqn: sedenion matrix eqn}
    \begin{bmatrix}
            y_0\\ y_1\\ y_2\\ y_3\\ \vdots\\ y_{12} \\ y_{13}\\ y_{14}\\ y_{15}
    \end{bmatrix} =
    \begin{bmatrix}
            w_0 &  -w_1 &  -w_2 &  -w_3 &  \ldots &  -w_{12} &  -w_{13} &  -w_{14} &  -w_{15} \\ 
          w_1 &  w_0 &  -w_3 &  w_2 & \ldots &  w_{13} &  -w_{12} &  -w_{15} &  w_{14} \\
          w_2 &  w_3 &  w_0 &  -w_1 &  \ldots &  w_{14} &  w_{15} &  -w_{12} &  -w_{13} \\
          w_3 &  -w_2 &  w_1 &  w_0 &  \ldots &  w_{15} &  -w_{14} &  w_{13} &  -w_{12} \\ 
          \vdots &  \vdots &  \vdots & \vdots &  \ddots &  \vdots &  \vdots &  \vdots & \vdots \\ 
          w_{12} &  -w_{13} &  -w_{14} &  -w_{15} &  \ldots &  w_0 &  w_1 &  w_2 &  w_3 \\ 
          w_{13} &  w_{12} &  -w_{15} &  w_{14} &  \ldots &  -w_1 &  w_0 &  -w_3 &  w_2 \\ 
          w_{14} &  w_{15} &  w_{12} &  -w_{13} &  \ldots &  -w_2 &  w_3 &  w_0 &  -w_1 \\ 
          w_{15} &  -w_{14} &  w_{13} &  w_{12} &  \ldots &  -w_3 &  -w_2 &  w_1 &  w_0 \\
    \end{bmatrix} 
    \begin{bmatrix}
            x_1\\ x_2\\ x_3\\ x_4\\ \vdots\\ x_{12} \\ x_{13}\\ x_{14}\\ x_{15}
    \end{bmatrix}
\end{equation}
}

\section{Methods}
\label{sec: methods}

\subsection{Input representation}
\label{sec: input}


The input to the model consists of two types of information sources, static and dynamic, supporting the use of the proposed multi-modal representation. The static input accounts for city-based variations as the model was trained using data from all three cities. The static input goes through a vector learning block, and is used as the real component of the aggregated sedenion input. 
The dynamic input consists of 12 frames  (Fig.
\ref{fig: data sequence}) accounting for the 12 previous time frames, and was used as 12 imaginary components of the sedenion. The remaining three imaginary components were set to zero (Fig. \ref{fig: sedenion divisions}b).

\subsection{Output representation}
\label{sec: output}
The expected six time frames of the prediction comes from the last layer of our model which equally gives a 16 components output (sedenion convolution output). The actual output comes from handpicking the exact location of the expected time frame in the output sedenion (Fig. \ref{fig: model architecture}). The network can easily be adapted up to 12 frames output for one hour predictions without any further adjustment to the output layer.

\subsection{Proposed model}
\label{sec: model}
\begin{figure}[htbp]
    \centering
    \includegraphics[width=1\columnwidth]{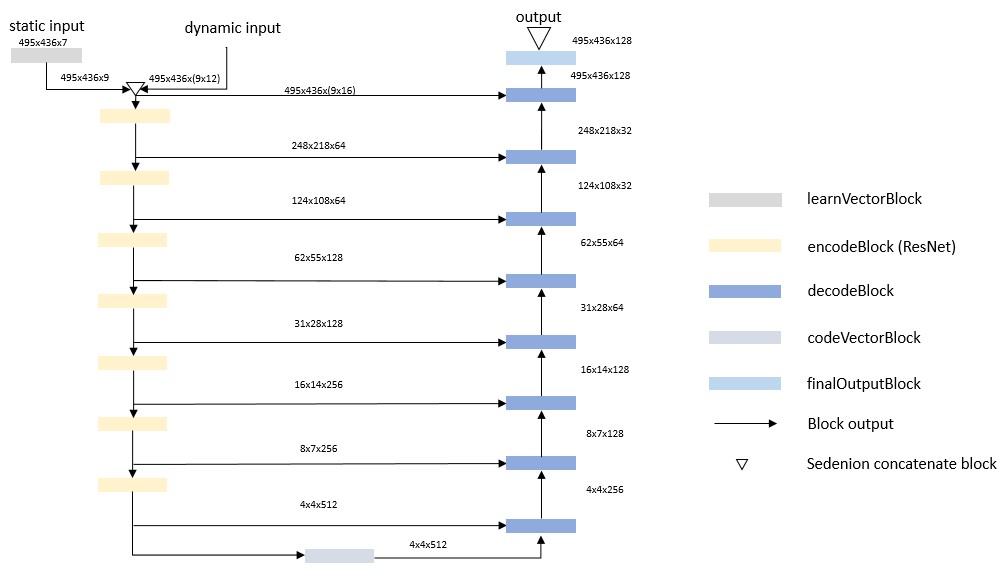}
    \caption{Overall model architecture.}
    \label{fig: model architecture}
\end{figure}

The proposed sedenion U-Net model structure is shown in Fig. \ref{fig: model architecture}. In this model, all convolutional blocks are sedenion convolutions except the ones in the learnVectorBlock, which helps in getting the static input (7 channels) into an equal channel dimension as the dynamic input (9 channels). 

\begin{figure}[htbp]
    \centering
    \includegraphics[width=1\columnwidth]{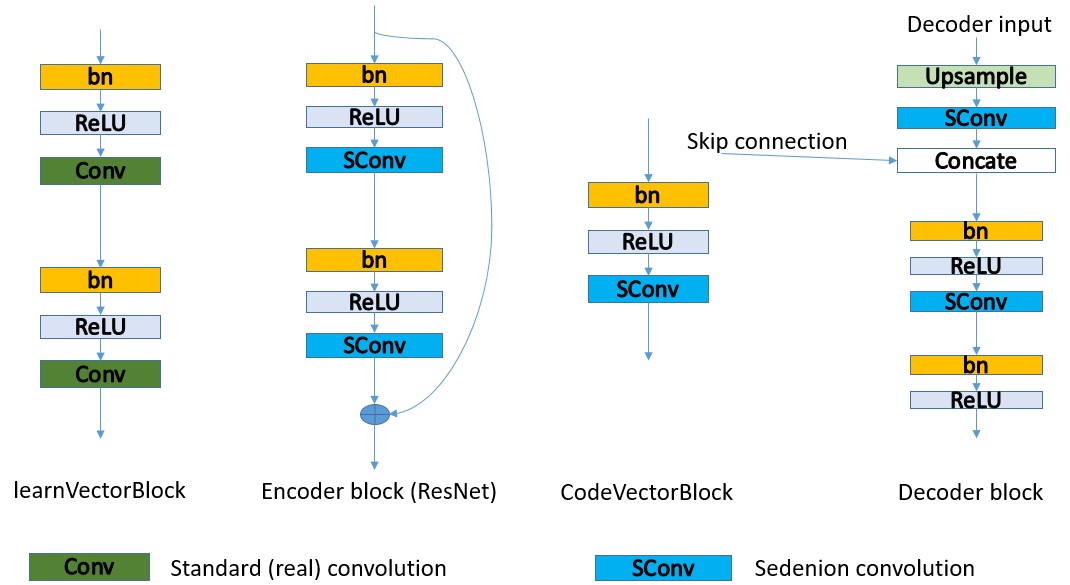}
    \caption{Model blocks}
    \label{fig: model blocks}
\end{figure}

The model blocks consist of various functions with their respective sub-blocks, shown in Fig. \ref{fig: model blocks}, as follows:

\begin{itemize}
    \item learnVectorBlock: This block serves as a feature extractor in case of any component having different number of dimension. It is used to extract a 9-channel feature from the 7-channel static input before concatenation with the dynamic input in the aggregated sedenion input of the network.
        
    \item encoderBlock: This is a ResNet block \cite{resnet}, which serves as the main encoder block for the U-Net model. Spatial pooling is done by strided convolution at the end of each encoder group.
    \item codeVectorBlock: This is a sedenion convolution preceded by batch normalization and ReLU.
    \item decoderBlock: This block has two inputs, one comes from the decoding end, while the other one from a skip connection, originating from the encoder output.
\end{itemize}

\section{Results}
\begin{figure}[tbp]
    \centering
    \includegraphics[width=0.6\columnwidth]{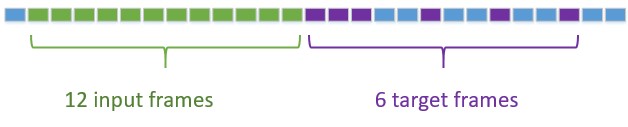}
    \caption{Data sequence}
    \label{fig: data sequence}
\end{figure}

In this competition, there are 362 day data files for each target city. 181 files were used for training, 18 for validation, and 163 were used for hold-out test set evaluation. The evaluation metric was the mean squared error. 
A single day training file contains 288 (24 hours x 60 minutes / 5 minute interval) time frame traffic map information. A sliding window was applied to extract 12 frame inputs and 6 frame outputs (Fig. \ref{fig: data sequence}).

The U-Net model described in Fig. \ref{fig: model architecture} was implemented in Pytorch \cite{NEURIPS2019_9015_pytorch}. The mean squared error (MSE) was used as the loss function, with the Adam optimizer \cite{Kingma2015AdamAM}. The learning rate was initially set to 1e-4 and was manually reduced to 1e-6 when performance plateaued on the validation set. 

The model was trained on a machine running two GeForce RTX 2080 Ti. The resources limitation and the high resolution of the data imposed constraints on the number of experiments, which, in turn, limited the competitiveness of the simulation results. The proposed model performance is summarized in Table \ref{tab: model result}.

\begin{table}[htbp]
  \caption{Model evaluation result}
  \label{tab: model result}
  \centering
  \begin{tabular}{lll}
    \toprule
    Parameters     & Validation MSE     & Test MSE \\
    \midrule
    628,592	    &   1.33893e-03	    &   1.30845e-03 \\
    \bottomrule
  \end{tabular}
\end{table}

\bibliographystyle{IEEEtran}
\bibliography{neurips}
\end{document}